\documentclass[letterpaper]{article} % DO NOT CHANGE THIS
\usepackage[draft]{aaai25}  % DO NOT CHANGE THIS
\usepackage{times}  % DO NOT CHANGE THIS
\usepackage{helvet}  % DO NOT CHANGE THIS
\usepackage{courier}  % DO NOT CHANGE THIS
\usepackage[hyphens]{url}  % DO NOT CHANGE THIS
\usepackage{graphicx} % DO NOT CHANGE THIS
\usepackage{natbib}  % DO NOT CHANGE THIS AND DO NOT ADD ANY OPTIONS TO IT
\usepackage{caption} % DO NOT CHANGE THIS AND DO NOT ADD ANY OPTIONS TO IT
\usepackage{algorithm}
\usepackage{algorithmic}

\usepackage{amssymb}
\usepackage{multirow}
\usepackage{subcaption}
\usepackage{graphicx}
\usepackage{booktabs}
\usepackage[table]{xcolor}
\usepackage{caption}
\usepackage{lipsum} % 测试用可删
\usepackage{booktabs}
\usepackage{amsmath}
\usepackage{pifont}

\usepackage{newfloat}
\usepackage{listings}
\DeclareCaptionStyle{ruled}{labelfont=normalfont,labelsep=colon,strut=off} % DO NOT CHANGE THIS
\floatstyle{ruled}
\newfloat{listing}{tb}{lst}{}
\floatname{listing}{Listing}
\title{Learning Active Perception via Self‑Evolving Preference Optimization for GUI Grounding}

\author{
    Wanfu Wang,
    Qipeng Huang,
    Guangquan Xue,
    Xiaobo Liang,
    Juntao Li \thanks{\quad Corresponding Author}
}
\affiliations{

    Soochow University \\
    Key Laboratory of Data Intelligence and Advanced Computing, Soochow University \\
    \texttt{ \{xbliang, ljt\}@suda.edu.cn, \{wfwang, qphuang, gqxue\}@stu.suda.edu.cn}
}

\usepackage{bibentry}
\begin{document}

\maketitle

\begin{abstract}
Vision Language Models (VLMs) have recently achieved significant progress in bridging visual perception and linguistic reasoning. 
Recently, OpenAI o3 model introduced a zoom-in search strategy that effectively elicits active perception capabilities in VLMs, improving downstream task performance.
However, enabling VLMs to reason effectively over appropriate image regions remains a core challenge in GUI grounding, particularly under high-resolution inputs and complex multi-element visual interactions.
In this work, we propose \textbf{LASER}, a self-evolving framework that progressively endows VLMs with multi-step perception capabilities, enabling precise coordinate prediction.
Specifically, our approach integrate Monte Carlo quality estimation with Intersection-over-Union (IoU)-based region quality evaluation to jointly encourage both accuracy and diversity in constructing high-quality preference data.
This combination explicitly guides the model to focus on instruction-relevant key regions while adaptively allocating reasoning steps based on task complexity.
Comprehensive experiments on the ScreenSpot Pro and ScreenSpot-v2 benchmarks demonstrate consistent performance gains, validating the effectiveness of our method.
Furthermore, when fine-tuned on GTA1-7B, LASER achieves a score of \textbf{55.7} on the ScreenSpot-Pro benchmark, establishing a \textbf{new state-of-the-art (SoTA)} among 7B-scale models.

\end{abstract}
\begin{links}
    \link{Code}{https://github.com/wwfnb/Laser}
    % \link{Datasets}{https://aaai.org/example/datasets}
    % \link{Extended version}{https://aaai.org/example/extended-version}
\end{links}
%%%%%%%%%%%%%%%%%%%%%%%%%%%%%%% Introduction %%%%%%%%%%%%%%%%%%%%%%%%%%%%%%%
\section{Introduction}

\begin{figure}[t]
\centering
\begin{subfigure}[t]{0.9\linewidth}
    \centering
    \includegraphics[width=\linewidth]{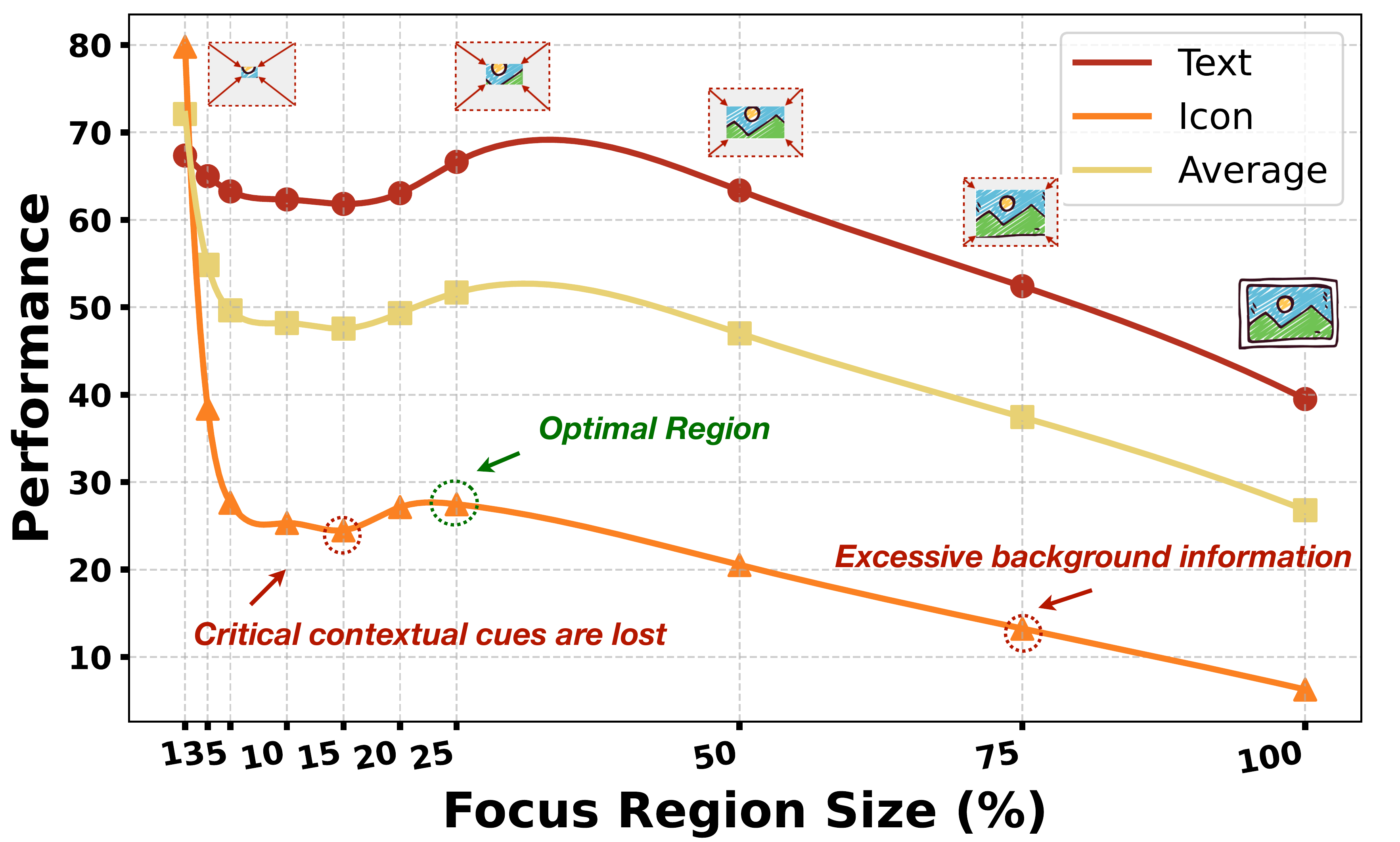} 
    \caption{Model performance is affected by varying the size of the ``\textbf{Focus Region}'' around the Ground Truth Bounding Box.}
    \label{fig:abstract_f1}
\end{subfigure}
\vspace{0.0em}  % 可选：控制上下间距
\begin{subfigure}[t]{0.9\linewidth}
    \centering
    \includegraphics[width=\linewidth]{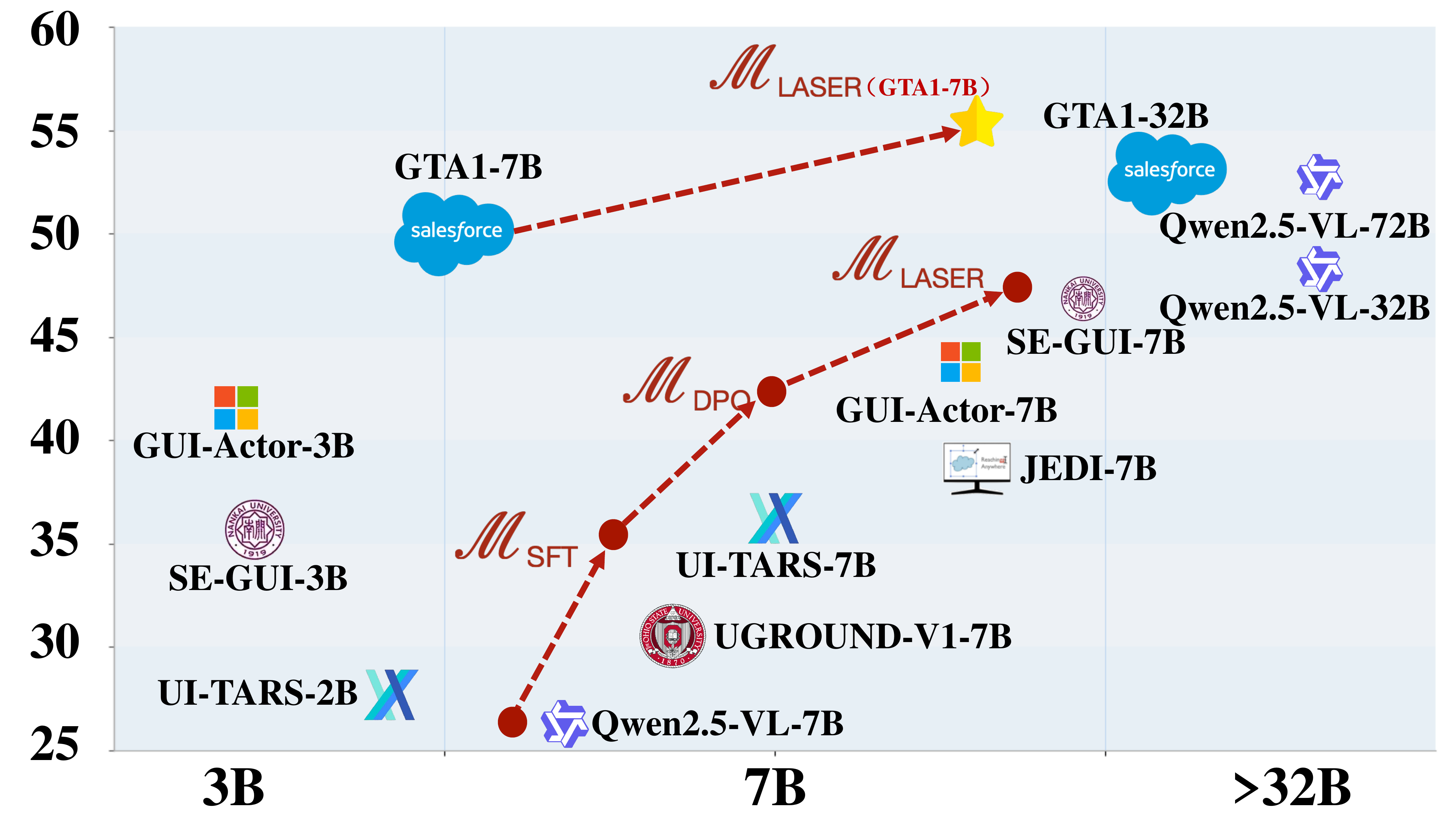} 
    \caption{The bottom Figure shows the performance curves of our \textbf{LASER} model across different training stages, achieving \textbf{SoTA} performance based on the GTA1-7B backbone.}
    \label{fig:abstract_f2}
\end{subfigure}
\label{fig:combined_results}
\end{figure}

Training autonomous agents to interact with graphical user interfaces (GUIs) presents unique challenges due to the high complexity of visual observations and the compositional nature of user interface    tasks~\cite{xi2025rise,qin2024tool}. 
However, the effectiveness of such agents can be largely attributed not only to their ability to ground language in visual observations, but also to their capability in active perception.
In particular, active perception~\cite{zhu2025active} is critical for guiding attention toward semantically relevant regions, thereby facilitating accurate action selection in complex GUI environments.
In GUI grounding tasks, most existing approaches adopt a direct prediction paradigm~\cite{qin2025ui,wu2024atlas,hong2024cogagent,xu2024aguvis}, where the model infers the target location and corresponding action in a single reasoning step.
Although widely used, this method often fails in \textbf{\textit{high-resolution}} or \textbf{\textit{multi-element interaction}} scenarios, primarily due to the absence of active perception capabilities.
Our preliminary analysis shows that the choice of visual focus region plays a pivotal role in the performance of VLMs (Qwen2.5-VL-7B). 
As illustrated in Figure~\ref{fig:abstract_f1}, explicitly focusing on appropriate regions yields substantial performance gains on the \textbf{ScreenSpot-Pro benchmark}~\cite{li2025screenspotpro}.
Such appropriate regions can be characterized as those that \textbf{\textit{(i) minimize background noise}} and (ii) \textbf{\textit{retain key contextual cues grounded in the user instruction}}, which are crucial for accurate model prediction.

Recent advances such as OpenAI’s o3~\cite{openai2025gpto3} model have introduced a novel paradigm termed ``\textbf{\textit{thinking with image}}''~\cite{su2025thinking}, in which intermediate visual transformations of the input are incorporated into the reasoning process as part of a visual chain-of-thought (CoT). 
This paradigm enables \textbf{\textit{a new axis of test-time scaling}}, effectively decomposing complex reasoning tasks into a sequence of simpler, more tractable subtasks.
Inspired by the success of long CoT strategies in language reasoning~\cite{chen2025towards}, this approach suggests that constructing visual CoT through multi-step reasoning can significantly enhance model performance on challenging visual tasks~\cite{zhu2025active,zhang2025chain,su2025pixel,zheng2025deepeyes}.
However, how to enable open-source VLMs to acquire such active perception capabilities remains an open and underexplored challenge, particularly in GUI grounding tasks.

In this work, we identify two core challenges in enabling active perception for GUI grounding:
(1) \textbf{\textit{how to evaluate preferences over candidate focus regions ?}} and 
(2) \textbf{\textit{how to allocate the model’s reasoning budget based on task difficulty ?}}
To address the first challenge, we introduce two complementary quality estimation techniques, Monte Carlo and IoU-based estimation, aimed at constructing preference data that jointly emphasizes both \textbf{accuracy} and \textbf{diversity}.
The Monte Carlo approach identifies focus regions that consistently lead to incorrect actions and treats them as negative examples, thereby improving the reliability of preference supervision. 
In contrast, the IoU-based method filters out preference pairs with high spatial overlap to ensure sufficient 
diversity among candidate regions.
To address the second challenge, we construct multi-step reasoning trajectories through an iterative refinement process, where the number of reasoning steps dynamically adapts to the complexity of the task. 
This mechanism enables the model to allocate its inference budget more effectively.
Building on these methods, we further propose \textbf{a self-evolving optimization framework, named LASER}, which enables the model to bootstrap its active perception capabilities through rejection sampling–based SFT and region-wise preference learning, without relying on extensive human supervision.

We conduct extensive experiments and analyses of LASER on two high-resolution GUI grounding benchmarks: ScreenSpot-Pro and ScreenSpot-v2.
As shown in Figure~\ref{fig:abstract_f2}, LASER trained with Qwen2.5-VL-7B achieves a score of \textbf{47.5} on ScreenSpot-Pro, surpassing the baseline by \textbf{+20.7} points and outperforming several RL–based methods, including GUI-R1 and SE-GUI.
Furthermore, when trained with GTA1-7B, LASER achieves a score of 55.7, even outperforming the much larger Qwen2.5-VL-72B, demonstrating both the effectiveness and scalability of our method.

In summary, our key contributions are as follows:
\begin{enumerate}
    \item We propose \textbf{LASER}, a self-evolving framework that performs multi-stage region-wise preference optimization, enabling VLMs to progressively focus on crucial regions and adaptively execute multi-step perception.
    \item The regions automatically predicted by LASER on the benchmark closely align with the prior distribution shown in Figure~\ref{fig:abstract_f1} (approximately \textbf{$20\%$} of the original image), which further validates the effectiveness of our region selection strategy.
    \item \textbf{LASER} trained with GTA1-7B achieves \textbf{a new SoTA score of 55.7} on the ScreenSpot-Pro benchmark.
\end{enumerate}

%%%%%%%%%%%%%%%%%%%%%%%%%%%%%%% Related Work %%%%%%%%%%%%%%%%%%%%%%%%%%%%%%%
\section{Related Work}

\subsection{Autonomous GUI Agents}
Training agents to assist humans in accomplishing complex tasks within GUI environments remains a significant challenge.
Current mainstream approaches can be categorized into \textit{textual-based agents}~\cite{deng2023mind2web, zhou2023webarena, lai2024autowebglm} and 
\textit{vision-based agents}~\cite{hong2024cogagent, qin2025ui, xu2024aguvis}, which rely on analyzing textual representations (such as HTML code or accessibility metadata) and visual inputs (such as interface screenshots), respectively.
In many real-world scenarios, textual representations fail to capture the spatial informations and handle complex GUI interface~\cite{huang2024understanding}. 

Prior work addresses this challenge through large-scale data synthesis~\cite{wu2024atlas,gou2024navigating,xie2025scaling}, involving collection of human instructions, GUI screenshots, and corresponding grounding coordinates, along with reinforcement learning~\cite{yang2025gta1, yuan2025enhancing, lu2025ui, luo2025gui} to encourage deep reasoning capabilities.
In contrast, we propose a self-evolving methods that encourages active perception behaviors in VLMs.
By progressively focusing on task-relevant regions, our approach guides the agent through step-by-step reasoning, thereby enabling more accurate and robust grounding.

\subsection{Thinking with Images}
OpenAI-o3~\cite{openai2025gpto3} is the first to introduce the ability to think with images into visual reasoning tasks by integrating visual information into chain-of-thought (CoT). This enables a new axis of test-time compute scaling, as the model actively transforms images by cropping, zooming, and rotating them. This interaction enables active perception by guiding the model to focus on task-relevant visual regions.
To stimulate this behavior, CoF~\cite{zhang2025chain} and Pixel-Reasoner~\cite{su2025pixel} distill trajectory data from strong model as a cold-start, then apply reinforcement learning to further encourage region-wise focus and reasoning. Recent approaches such as ACTIVE-o3~\cite{zhu2025active}, DeepEyes~\cite{zheng2025deepeyes}, and Ground-R1~\cite{cao2025ground} directly leverage reinforcement learning to train agents from scratch, requiring extensive exploration to obtain high-confidence action trajectories.
In contrast, our method introduces a self-evolving mechanism that progressively induces active perception behavior within the model itself. Notably, this capability emerges natively, without relying on external experts or imitation learning.

% 方法的overview图
\begin{figure*}[t]
\centering
\includegraphics[width=1.0\textwidth]{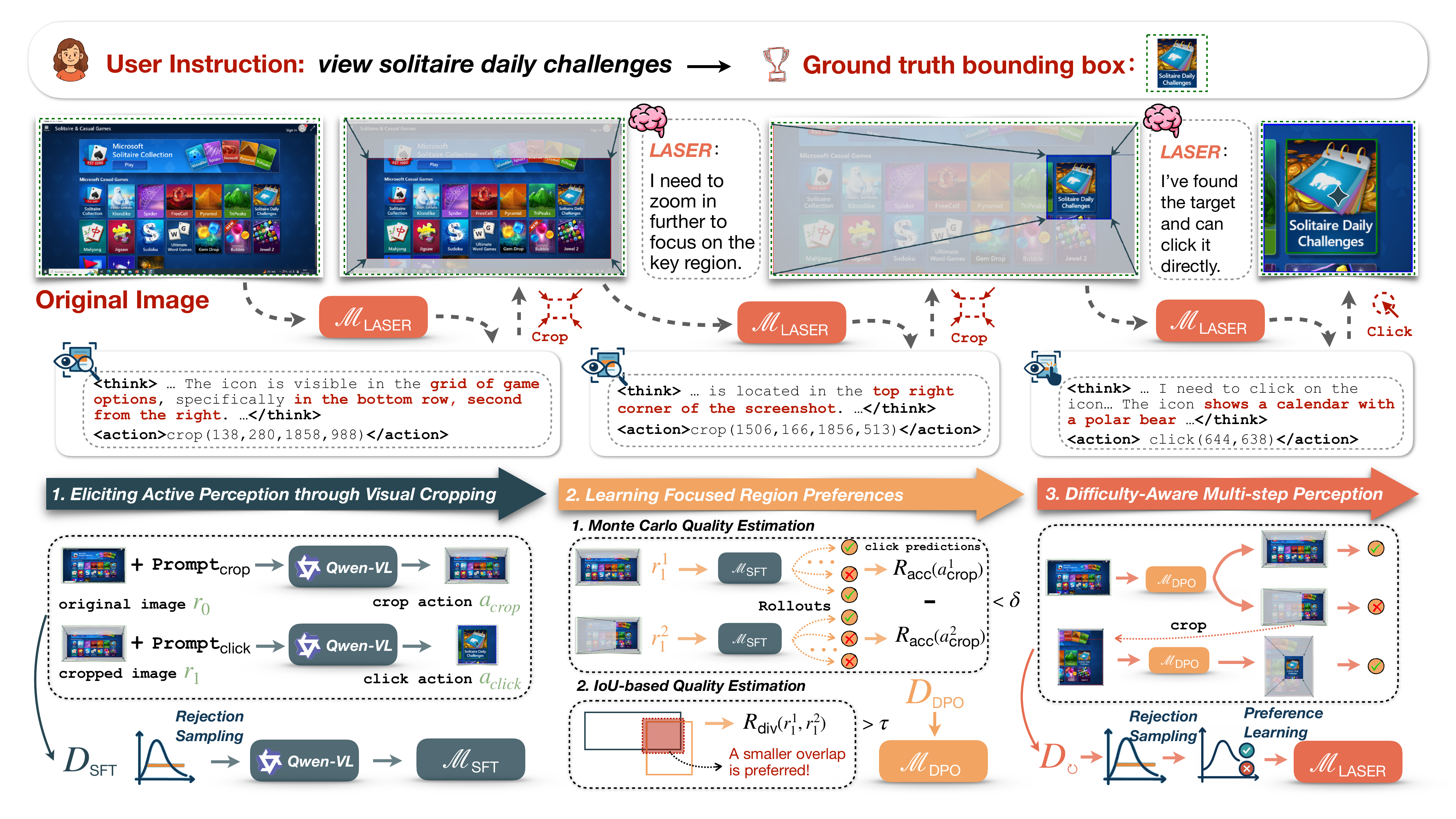} 
\caption{Overview of the proposed LASER framework. 
Given a user instruction and the original image, the trained $\mathcal{M}_{\text{LASER}}$ model progressively focuses on key regions through a multi-step reasoning process. 
At each step, the Visual CoT captures critical cues (highlighted in red within the $\texttt{<think>}$ tag) based on the current focus region.
Below, we also illustrate the multi-stage self-evolving optimization process that elicits LASER’s multi-step active perception capabilities.}
\label{fig:method_overview}
\end{figure*}

\section{Problem Statement}

The goal of GUI grounding is to identify and localize the target region or interface element in response to a user query. 
While recent approaches often employ VLMs to directly predict the target location, the lack of explicit reasoning limits their capacity for test-time scaling.
In this work, we introduce an auxiliary ``\textbf{focus region}'', defined as an instruction-relevant area within the original GUI image, to enable multi-step visual reasoning.
At each time step $t$, the model $\mathcal{M}$ receives an image (either the original screenshot $r_0$ or a previously cropped region $r_t$) along with the natural language instruction, and predicts the next action $a_t$. The action space includes two options:
\begin{itemize}
    \item $\mathcal{A}_{crop}$: a rectangular bounding box predicted by $\mathcal{M}$ that defines the next region $\text{Crop} (x, y, w, h)$ within $I_t$.
    \item $\mathcal{A}_{click}$: the coordinate of the final click location $\text{Click} (x, y)$ corresponding to the user instruction.
\end{itemize}
We aim to optimize a policy $\pi_\theta$ with active perception capabilities, such that:
\begin{equation}
    \theta^{*} = \arg \max_{\theta} \mathbb{E}{ \left[  \sum^{T}_{t=0} \log \pi_{\theta}(a_t | s_{t}) \right] }, s_t = [I, s_{t-1}],
\end{equation}
where $s_t$ denotes the current state, comprising the instruction $I$, the cropped image $r_t$, and the trajectory history up to step $t-1$.
Our hypothesis is that such perceptual behavior can \textbf{emerge through self-evolving learning}, without relying on external supervision or handcrafted labels.
However, a core challenge remains: \textbf{\textit{how to evaluate the quality of the selected focus regions ?}}

%%%%%%%%%%%%%%%%%%%%%%%%%%%%%%% Method %%%%%%%%%%%%%%%%%%%%%%%%%%%%%%%

\section{Method}

\subsection{Overview}
Our approach operates with three main steps, progressively eliciting multi-step active perception capabilities (Figure~\ref{fig:method_overview}):
\begin{enumerate}
    \item \textbf{\textit{Eliciting Active Perception through Visual Cropping}}: 
    Given the paired training data, we prompt the VLM backbone $\mathcal{M}_{\text{RAW}}$ to predict a focused region. 
    The corresponding region is then cropped from the original image and integrated into the CoT as visual context, guiding the model toward accurate click-coordinate prediction.
    To improve the quality of reasoning trajectories, we adopt a STaR-style rejection sampling strategy to construct the dataset $\mathcal{D}_{\text{SFT}}$, which is used to finetune $\mathcal{M}_{\text{SFT}}$.
    \item \textbf{\textit{Learning Focused Region Preferences}}: 
    We sample multiple reasoning trajectories from $\mathcal{M}_{\text{SFT}}$ and estimate region-wise preferences using Monte Carlo Estimation.
    An IoU-based filter is applied to remove low-quality candidates. 
    The resulting preference pairs dataset $\mathbf{ \mathcal{D}_{\text{DPO}} }$ are used to train a stronger model $\mathcal{M}_{\text{DPO}}$ via DPO.
    \item \textbf{\textit{Difficulty-Aware Multi-step Perception}}:
    While $\mathcal{M}_{\text{DPO}}$ supports single-step perception, it is prone to failure in complex scenarios that demand deeper reasoning.
    To overcome this limitation, we allow $\mathcal{M}_{\text{DPO}}$ to iteratively generate multi-step reasoning trajectories, enabling the construction of a diverse and difficulty-aware training data. 
    The final model is then trained on this multi-step dataset $\mathbf{ \mathcal{D}_{\circlearrowright} }$, making it with the ability to dynamically adjust reasoning depth based on the difficulty of the query.
\end{enumerate}

\subsection{Eliciting Active Perception}\label{sec:elicition_active_perception}
While it is feasible to instruct $\mathcal{M}_{\text{RAW}}$ to perform perception tasks directly via task prompts, this process is often unstable and yields low accuracy.
To address this issue, we synthesize high-quality training data via rejection sampling to explicitly guide the model toward learning the desired reasoning behavior.
Formally, given the original screenshot $r_0$ and user instruction $I$, the model first predicts a focused region: $a_{crop} = \mathcal{M}_{\text{RAW}}(\texttt{Prompt}_{\text{crop}}, r_0)$, where $\texttt{Prompt}_{\text{Crop}}$ refers to the instruction formatted to elicit a cropping action~\footnote{For simplicity, we do not explicitly denote the nature language CoT associated with the cropping action $a_{\text{crop}}$.}. 
The resulting cropped region $r_1$ is then used as input for the second step, where the model is prompted to predict the final click location:
$a_{click} = \mathcal{M}_{\text{RAW}}(\texttt{Prompt}_{\text{click}}, r_1)$.

To ensure the quality of the synthesized data, we filter out incorrect predictions by comparing the model-generated click location with the ground-truth click label $\hat{a}_{click}$ provided in the dataset.
Each cropping action $a_{\text{crop}}$ is represented by a tuple of two coordinates $\text{Crop} (x, y, w, h)$ that defines the focused region.
We then organize the all reasoning trajectory $\mathcal{D}_{\text{SFT}} = \{(r_0, I, a_{crop}, r_1, a_{click})_{k=0}^{|D_{\text{SFT}}|} \}$~\footnote{Note that $a_{\text{click}}$ denotes the model predicted result, which may differ from the ground-truth label $\hat{a}_{\text{click}}$.} into a multi-turn conversational format, which is used to supervise the model $\mathcal{M}_{\text{SFT}}$ through supervised fine-tuning.

\subsection{Region-wise Preference Learning}
\label{sec:region_wise_preference_learning}
However, $\mathcal{M}_{\text{SFT}}$ merely learns to mimic the behavior of the rejected sampling distribution and still lacks true perceptual capability.
One of the most critical failure cases the predicted focus region $r_1$ fails to include the ground-truth bounding box.
To address this, we aim to teach the model to distinguish between focused and unfocused regions in a human-like manner through preference learning.
Nevertheless, even for humans, it is non-trivial to define a precise evaluation metric to compare different focus regions.
So, we design two methods for assigning preference scores to different reasoning trajectories:
\begin{enumerate}
    \item \textbf{Monte Carlo Quality Estimation}: we estimate the quality of a perception trajectory by computing the success rate of subsequent inference steps, inspired by Math-Shepherd~\cite{wang2023math}.
    Formally, given a pair of perception action $(a^{1}_{\text{crop}}, a^{2}_{\text{crop}})$, we define \textbf{accuracy reward $\mathcal{R}_{acc}$} of their quality by measuring the proportion of crop action leads to the correct click predictions:
    \begin{align}
        \mathcal{R}_{acc} (a^{1}_{\text{crop}}, a^{2}_{\text{crop}}) &= \frac{1}{N} \sum_{i=1}^N \mathbb{I} (a^{(i)}_{\text{click}} = \hat{a}_{\text{click}}), \\ 
        a^{(i)}_{\text{click}} &= \pi_{\theta} (a | r_0, a_{\text{crop}}, r_1, I),  \nonumber 
    \end{align}
    where $a^{(i)}_{\text{click}}$ denotes the predicted click action from the $i$-th rollout based on $a_{\text{crop}}$, and $\pi_{\theta}$ is the policy of $\mathcal{M}_{\text{SFT}}$.
    \item \textbf{IoU-based Quality Estimation}: we further introduce the \textbf{diversity reward $\mathcal{R}_{\text{div}}$} by computing the Intersection over Union (IoU) between the cropped regions $r_1^1$ and $r_1^2$ obtained from two different crop actions $(a^{1}_{\text{crop}}, a^{2}_{\text{crop}})$:
    \begin{align}
        \mathcal{R}_{\text{div}} (r_1^1, r_1^2) = \text{IoU} (r_1^1, r_1^2) =  \frac{|r_1^1 \cap r_1^2|}{|r_1^1 \cup r_1^2|}
    \end{align}
    where $\text{IoU}(r_1^1, r_1^2)$ measures the normalized overlap area between the two regions. 
    A smaller $\mathcal{R}_{\text{div}}$ encourages exploration of complementary focus regions, which is beneficial in ambiguous or multi-target scenarios.
\end{enumerate}
Based on the above reward definitions, we further employ model $\mathcal{M}_{\text{SFT}}$ to sample multiple pairs of perception trajectories and filter them using both rewards to construct a high-quality preference dataset.
Specifically, we introduce two threshold hyper-parameters, $\delta$ and $\tau$, to jointly control the correctness and diversity of the sampled data:

\begin{align}
    \mathcal{D}_{\text{DPO}} &= \big\{ (r_0, I, a^{1}_{\text{crop}}, a^{2}_{\text{crop}})_{k=0}^{|D_{\text{DPO}}|} \ \big| \\ 
    & \mathcal{R}_{\text{acc}} (a^{1}_{\text{crop}}, a^{2}_{\text{crop}}) > \delta,
    \mathcal{R}_{\text{div}} (r_1^1, r_1^2) < \tau \big\}. \nonumber
\end{align}
This ensures that the selected pairs not only lead to correct subsequent actions with high probability but also demonstrate sufficient perceptual diversity, which is critical for preference learning.
We then leverage the filtered dataset $\mathcal{D}_{\text{DPO}}$ to train the model $\mathcal{M}_{\text{DPO}}$ via DPO method.

\subsection{Multi-step Perception}
\label{sec:multi_step_perception}
In high-resolution or complex scenarios, visual reasoning often requires multi-step perception, where the model progressively focuses on key regions to complete the reasoning process. 
Ideally, the model should adaptively allocate reasoning cost according to instruction difficulty. 
However, the original $\mathcal{M}_{\text{DPO}}$ only supports single-step perception and cannot directly synthesize or optimize multi-step trajectories.
To address this limitation, we take $\mathcal{M}_{\text{DPO}}$ as a context-independent iterative reasoning model, where each iteration takes the cropped region $r_t$ as the original image and predict the subsequent crop action. 
This formulation enables iterative perception without requiring architectural modifications or additional retraining.

Specifically, we collect single-step failure cases in which the predicted crop region contains the ground-truth bounding box but results in an incorrect click action.
For these cases, we iteratively apply $\mathcal{M}_{\text{DPO}}$ to perform multi-step reasoning, and include the trajectory in a new dataset $\mathbf{ \mathcal{D}_{\circlearrowright} }$ if the final step successfully corrects the click action.
Following the same training strategy as before, we apply rejection sampling for SFT and adopt region-wise DPO to fine-tune the multi-step policy, yielding our final $\mathcal{M}_{\text{LASER}}$ model.

%%%%%%%%%%%%%%%%%%%%%%%%%%%%%%% Experiments （薛广泉先写）%%%%%%%%%%%%%%%%%%%%%%%%%%%%%%%

%%%%%%%%%%%%%%%%%%%%%%%%%%%%%%%%%%%%%%%%%%%%%%%%%%%
%%%%%%%%%%%%%%%%% 主试验的表 - ScreenSpot-Pro        
%%%%%%%%%%%%%%%%%%%%%%%%%%%%%%%%%%%%%%%%%%%%%%%%%%%
\begin{table*}[t]
\centering
\small
\setlength{\tabcolsep}{1.8mm}  % ← 设置列间距为1mm
\begin{tabular}{l|cc|cc|cc|cc|cc|cc|cc ccc}
\toprule
\multirow{2}{*}{\textbf{Model}} & \multicolumn{2}{c|}{\textbf{CAD}} & \multicolumn{2}{c|}{\textbf{Dev}} & \multicolumn{2}{c|}{\textbf{Creative}} & \multicolumn{2}{c|}{\textbf{Scientific}} & \multicolumn{2}{c|}{\textbf{Office}} & \multicolumn{2}{c|}{\textbf{OS}} & \multicolumn{3}{c}{\textbf{Avg.}} \\
 & Text & Icon & Text & Icon & Text & Icon & Text & Icon & Text & Icon & Text & Icon & Text & Icon & Avg. \\
\midrule
\rowcolor{gray!20}
\multicolumn{16}{l}{\textit{Proprietary Models}} \\
GPT-4o & 2.0 & 0.0 & 1.3 & 0.0 & 1.0 & 0.0 & 2.1 & 0.0 & 1.1 & 0.0 & 0.0 & 0.0 & 1.3 & 0.0 & 0.8 \\
Claude & 14.5 & 3.7 & 22.0 & 3.9 & 25.9 & 3.4 & 33.9 & 15.8 & 30.1 & 16.3 & 11.0 & 4.5 & 23.4 & 7.1 & 17.1 \\
\midrule
\rowcolor{gray!20}
\multicolumn{16}{l}{\textit{General Open-source Models}} \\
Qwen2.5-VL-3B & 9.1 & 7.3 & 22.1 & 1.4 & 26.8 & 2.1 & 38.2 & 7.3 & 33.9 & 15.1 & 10.3 & 1.1 & 23.6 & 3.8 & 16.1 \\
Qwen2.5-VL-7B & 16.8 & 1.6 & 46.8 & 4.1 & 35.9 & 7.7 & 49.3 & 7.3 & 52.5 & 20.8 & 37.4 & 6.7 & 38.9 & 7.1 & 26.8 \\
% Qwen2.5-VL-72B & -- & -- & -- & -- & -- & -- & -- & -- & -- & -- & -- & -- & -- & -- & 53.3 \\
% Kimi-VL-A3B & -- & -- & -- & -- & -- & -- & -- & -- & -- & -- & -- & -- & -- & -- & 51.0 \\
% Kimi-VL-A3B-Thinking-2506 & -- & -- & -- & -- & -- & -- & -- & -- & -- & -- & -- & -- & -- & -- & 51.0 \\
\midrule
% \multicolumn{16}{l}{\textit{SFT-based Models from jedi article}} \\
% SFT-Based
\rowcolor{gray!20}
\multicolumn{16}{l}{\textit{SFT-based Models}} \\
OS-Atlas-7B & 12.2 & 4.7 & 33.1 & 1.4 & 28.8 & 3.2 & 37.5 & 7.3 & 33.9 & 5.7 & 27.1 & 4.5 & 28.1 & 4.0 & 18.9 \\
UGround-7B & 14.2 & 1.6 & 26.6 & 2.1 & 27.3 & 2.8 & 31.9 & 2.7 & 31.6 & 11.3 & 17.8 & 0.0 & 25.0 & 2.8 & 16.5 \\
UGround-V1-7B & 15.8 & 1.2 & 51.9 & 2.8 & 47.5 & 9.7 & 57.6 & 14.5 & 60.5 & 13.2 & 38.3 & 7.9 & 45.2 & 8.1 & 31.1 \\
% UI-TARS-2B & 17.8 & 4.7 & 47.4 & 4.1 & 42.9 & 6.3 & 56.9 & 17.3 & 50.3 & 17.0 & 21.5 & 5.6 & 39.6 & 8.4 & 27.7 \\
UI-TARS-7B & 20.8 & 9.4 & 58.4 & 12.4 & 50.0 & 9.1 & 63.9 & \textbf{31.8} & 63.3 & 20.8 & 30.8 & 16.9 & 47.8 & 16.2 & 35.7 \\
UI-TARS-72B & 18.8 & 12.5 & 62.9 & 17.2 & 57.1 & 15.4 & 64.6 & 20.9 & 63.3 & 26.4 & 42.1 & 15.7 & 50.9 & 17.6 & 38.1 \\ 
% JEDI-3B & 27.4 & 9.4 & 61.0 & 13.8 & 53.5 & 8.4 & 54.2 & 18.2 & 64.4 & 32.1 & 38.3 & 9.0 & 49.8 & 13.7 & 36.1 \\
JEDI-7B & 38.0 & 14.1 & 42.9 & 11.0 & 50.0 & 11.9 & 72.9 & 25.5 & 75.1 & \underline{47.2} & 33.6 & 16.9 & 52.6 & 18.2 & 39.5 \\
\midrule
% \multicolumn{16}{l}{\textit{RL-based Models above come from SE-GUI}} \\
% RL-Based
\rowcolor{gray!20}
\multicolumn{16}{l}{\textit{RL-based Models}} \\
% UI-R1-3B & 11.2 & 6.3 & 22.7 & 4.1 & 27.3 & 3.5 & 42.4 & 11.8 & 32.2 & 11.3 & 13.1 & 4.5 & 24.9 & 6.4 & 17.8 \\
% GUI-R1-3B & 26.4 & 7.8 & 38.8 & 4.8 & 40.9 & 5.6 & 61.8 & 17.3 & 53.6 & 17.0 & 28.1 & 5.6 & -- & -- & -- \\
GUI-R1-7B & 23.9 & 6.3 & 49.4 & 4.8 & 38.9 & 5.4 & 55.6 & 11.8 & 58.7 & 26.4 & 42.1 & 16.9 & -- & -- & -- \\
% SE-GUI-3B & 38.1 & 12.5 & 55.8 & 7.6 & 47.0 & 4.9 & 61.8 & 16.4 & 59.9 & 24.5 & 40.2 & 12.4 & 50.4 & 11.8 & 35.9 \\
SE-GUI-7B & 51.3 & \textbf{42.2} & 68.2 & 19.3 & 57.6 & 9.1 & 75.0 & 28.2 & 78.5 & 43.4 & 49.5 & 25.8 & 63.5 & 21.0 & 47.3 \\
GTA1-7B  & \underline{53.3} & 17.2 & 66.9 & 20.7 & 62.6 & 18.2  & \underline{76.4} & \textbf{31.8} & \textbf{82.5} & \textbf{50.9} & 48.6 & 25.9 & 65.5 & 25.2 & 50.1 \\
GTA1-32B & 43.7 & 23.4 & \textbf{82.5} & \underline{28.3} & \textbf{69.2} & 14.7 & \textbf{79.9} & \textbf{31.8} &  \underline{80.8} & 43.4 & \textbf{70.1} & \underline{32.6} & \underline{69.9}&  \underline{27.2} & \underline{53.6} \\
\midrule
\rowcolor{gray!20}
\multicolumn{16}{l}{\textit{Ours (7B Backbone)}} \\
% \textbf{Self-Evovle-3B} & 38.1 & 12.5 & 55.8 & 7.6 & 47.0 & 4.9 & 61.8 & 16.4 & 59.9 & 24.5 & 40.2 & 12.4 & 50.4 & 11.8 & 35.9 \\
% \textbf{Self-Evol-Qwen2.5-VL-7B-sft} & \underline{38.5} & 7.8 & \textbf{70.7} & \textbf{24.1} & \textbf{58.5} & \textbf{23.7} & 64.5 & 21.8 & 69.4 & 33.9 & \textbf{61.6} & \underline{23.5} & \underline{59.6} & \textbf{22.6} & \underline{45.5} \\
% \textbf{Self-Evol-Qwen2.5-VL-7B-dpo} & \underline{48.7} & 23.4 & \textbf{70.7} & \textbf{24.1} & \textbf{57.5} & \textbf{24.4} & 68.7 & 21.8 & 71.7 & 35.8 & \textbf{54.2} & \underline{23.5} & \underline{61.7} & \textbf{24.6} & \underline{47.5} \\
\textbf{LASER (Qwen2.5-VL)} & 48.7 & 23.4 & 70.7 & 24.1 & 57.5 & \textbf{24.4} & 68.7 & 21.8 & 71.7 & 35.8 & 54.2 & 23.5 & 61.7 & 24.6 & 47.5 \\
% \textbf{LASER-Qwen2.5-VL-7B} & \underline{48.7} & 23.4 & \textbf{70.7} & \textbf{24.1} & \textbf{57.5} & \textbf{24.4} & 68.7 & 21.8 & 71.7 & 35.8 & \textbf{54.2} & \underline{23.5} & \underline{61.7} & \textbf{24.6} & \underline{47.5} \\
% \textbf{Self-Evol-GTA1-7B} & \underline{67.0} & 31.2 & \textbf{79.8} & \textbf{28.9} & \textbf{66.1} & \textbf{20.9} & 75.6 & 30.0 & 79.6 & 50.9 & \textbf{57.0} & \underline{37} & \underline{71.3} & \textbf{30.6} & \underline{55.7} \\
\textbf{LASER (GTA1)} & \textbf{67.0} & \underline{31.2} & \underline{79.8} & \textbf{28.9} & \underline{66.1} & \underline{20.9} & 75.6 & \underline{30.0} & 79.6 & \textbf{50.9} & \underline{57.0} & \textbf{37} & \textbf{71.3} & \textbf{30.6} & \textbf{55.7} \\
\bottomrule
\end{tabular}
\caption{Performance comparison of different models on ScreenSpot-Pro. Results marked in \textbf{bold} represent the best performance, and those \underline{underlined} indicate the second-best performance.}
\label{tab:table_for_screen_spot_pro}
\end{table*}

\section{Experiments}
\subsection{Implementation Details}

\subsubsection{Training Details}

We use the training dataset from GTA-1~\cite{yang2025gta1}, which removes noisy samples from Aria-UI~\cite{yang2024aria} and OS-Atlas~\cite{wu2024atlas} using a set of heuristic rules. we randomly sample 270K instances as the initial data for our training pipeline.
We evaluate our model on two widely used GUI grounding benchmarks: \textbf{ScreenSpot-v2}~\cite{cheng2024seeclick} and \textbf{ScreenSpot-Pro}~\cite{li2025screenspotpro}, both designed to assess the grounding accuracy of GUI agents.
We use Qwen2.5-VL-7B as the base model, and all experiments are conducted on 8$\times$NVIDIA A100-40GB GPUs, ensuring reproducibility.
Following the procedures described in Section~\ref{sec:elicition_active_perception} and Section~\ref{sec:multi_step_perception}, we first synthesize 120K single-step trajectory samples to construct the supervised finetuning dataset $\mathcal{D}_{\text{SFT}}$, aiming to equip the model with single-step perception capabilities. To further improve the model's ability for multi-step perception, we synthesize 36K multi-step training samples (including both one-step and two-step trajectories), denoted as $\mathcal{D}_{\circlearrowright}$, to gradually guide the model toward active visual perception.For supervised finetuning, we use a batch size of 128 and a learning rate of $1 \times 10^{-5}$. The maximum resolution of each image is limited to 2,408,448 pixels. Training is conducted for a single epoch using the Adam optimizer.

During the DPO training stage, we construct 26K preference pairs from the single-step setting as $\mathcal{D}_{\text{DPO}}$, and 8K preference pairs in the multi-step setting for the preference optimization. To ensure data quality, we apply the \textit{score margin filter} with threshold $\delta = 4$ and the \textit{IoU overlap filter} with threshold $\tau = 0.8$. We adopt the sigmoid-based preference loss with a preference scaling factor $\beta = 0.1$, using a batch size of 128 and a learning rate of $3 \times 10^{-6}$.

\subsubsection{Baselines}
We benchmark our approach against a broad spectrum of leading models, spanning proprietary systems, general open-source VLMs, and task-specific agents.
Among proprietary solutions we include GPT-4o~\citep{hurst2024gpt} and the Claude Computer Use suite~\citep{anthropic2024computer}.
Representative open-source vision–language models are Qwen2.5-VL~\citep{bai2025qwen2} and Kimi-VL-A3B~\citep{team2025kimi}.
Supervised-fine-tuned baselines comprise OS-Atlas~\citep{wu2024atlas}, JEDI~\citep{xie2025scaling}, UGround~\citep{gou2024navigating}, and UI-TARS~\citep{qin2025ui}.
Reinforcement-learning-based competitors include SE-GUI~\citep{yuan2025enhancing}, GTA1~\citep{yang2025gta1}, UI-R1~\citep{lu2025ui}, and GUI-R1~\citep{luo2025gui}.
\subsection{Experimental Results}
As shown in Table~\ref{tab:table_for_screen_spot_v2} and Table~\ref{tab:table_for_screen_spot_pro}, we conduct comprehensive comparisons on both \textit{ScreenSpot-v2} and \textit{ScreenSpot-Pro} benchmarks. The evaluation covers six GUI domains and two task types (Text and Icon grounding). Our method, LASER, consistently outperforms previous models in terms of both overall grounding accuracy and generalization ability across different domains, demonstrating the effectiveness and robustness of our self-evolving training strategy.

Specifically, LASER (Qwen2.5-VL) , trained through our full self-evolving pipeline comprising multi-stage training and preference optimization achieves an average grounding accuracy of 47.5\% scores, outperforming its base model Qwen2.5-VL-7B (26.8\%) by a relative margin of 20.7\%. Compared to JEDI-7B (39.5\%) and SE-GUI-7B (47.3\%), LASER also shows a clear advantage validating the effectiveness of our self-evolving training strategy.
Notably, by fine-tuning GTA1-7B with our proposed dataset $\mathbf{ \mathcal{D}_{\circlearrowright} }$, LASER (GTA1) achieves a score of 55.7\%, surpassing the strongest GTAI-32B (53.6\%) which represents a relative improvement of 2.1\%, even outperforming the much larger Qwen2.5-VL-72B (53.3\%), which highlights the effectiveness of our training data.
These results highlight the dual strengths of our approach: (1) it enables small models to acquire strong GUI grounding capabilities through self-evolution; and (2) the synthetic trajectories can effectively elicit active perception capability of VLMs, resulting in substantial performance gains.

%%%%%%%%%%%%%%%%%%%%%%%%%%%%%%%%%%%%%%%%%%%%%%%%%%%
%%%%%%%%%%%%%%%%% 其他的表 - ScreenSpot-v2        
%%%%%%%%%%%%%%%%%%%%%%%%%%%%%%%%%%%%%%%%%%%%%%%%%%%
\begin{table}[t]
\centering
\small
\setlength{\tabcolsep}{1mm}  % ← 设置列间距为1mm
% \begin{tabular}{l|cc|cc|cc|cc|ccc}
\begin{tabular}{l|cc|cc|cc|c}
\toprule
\multirow{2}{*}{\textbf{Model}} & \multicolumn{2}{c|}{\textbf{Desktop}} & \multicolumn{2}{c|}{\textbf{Mobile}} & \multicolumn{2}{c|}{\textbf{Web}} & \multirow{2}{*}{\textbf{Avg}} \\
 % & Text & Icon & Text & Icon & Text & Icon & Text & Icon & Avg. \\
 & Text & Icon & Text & Icon & Text & Icon \\
\midrule
\rowcolor{gray!20}
\multicolumn{8}{l}{\textit{General Open-source Models}} \\
SeeClick &70.1 &29.3 &78.4 &50.7 &55.2 &32.5  & 55.1 \\
OS-Atlas-Base-7B  & 89.7 & 69.3 & 96.2 & 83.4 &\underline{94.0} &79.8  &87.1 \\
Qwen2.5-VL-7B  & 90.2 & 74.2 & 97.6 & 87.2  & 93.2 & 81.3   & 88.8 \\
UI-TARS-7B  & 95.4 & 85.0 & 96.9 & \underline{89.1} & 93.6 & \underline{85.2} & 91.6 \\
\midrule
\rowcolor{gray!20}
\multicolumn{8}{l}{\textit{RL-based Models}} \\
SE-GUI-7B & -- & -- & -- & -- & -- & --  & 90.3 \\
GTA1-7B & 94.9 & \textbf{89.3} & \textbf{99.9} & 88.6 & 92.3 & \textbf{86.7}  & \textbf{92.4} \\
\midrule
\rowcolor{gray!20}
\multicolumn{8}{l}{\textit{SFT-based Models}} \\
JEDI-3B & \underline{96.9} & 78.6 & 96.6 & 81.5 & 88.5 & 83.7  & 88.6  \\  
JEDI-7B & 95.9 & \underline{87.9} & 96.9 & 87.2 & \textbf{94.4} & 84.2  & \underline{91.7}
 % \\
% OS-Atlas-Base-7B & 90.2 & \textbf{66.4} & \underline{93.8} & \underline{79.9} & \textbf{92.6} & \textbf{79.1}  & \textbf{85.1}
 \\
\midrule
\rowcolor{gray!20}
\multicolumn{8}{l}{\textit{Ours}} \\
% \textbf{Self-Evovle-3B} & 38.1 & 12.5 & 55.8 & 7.6 & 47.0 & 4.9 & 61.8 & 16.4 & 59.9 & 24.5 & 40.2 & 12.4 & 50.4 & 11.8 & 35.9 \\
% \textbf{Self-Evovle-7B-sft} & 96.9 & 80.0 & \textbf{98.9} & \textbf{90.5} & \underline{93.5} & 82.2  & 
% \underline{91.5} \\
\textbf{LASER (Qwen2.5-VL)} & 93.3 & 75.0 & \underline{98.6} & 88.2 & 93.2 & 81.3  & 89.7 \\
\textbf{LASER (GTA1)} & \textbf{97.4} & 87.1 & 97.9 & \textbf{89.6} & 93.2 & \underline{85.2}  & \textbf{92.4} \\
\bottomrule
\end{tabular}
\caption{Performance comparison of various models on the \textbf{ScreenSpot-v2} benchmark.
}
\label{tab:table_for_screen_spot_v2}
\end{table}

\section{Discussion}

%%%%%%%%%%%%%%%% 过滤规则 IOU 和 Score margin的 消融试验表 %%
\begin{table}[t]
\centering
\begin{tabular}{c c c c c c}
    \toprule
     & $\mathcal{R}_{\text{div}}$ & $\mathcal{R}_{\text{acc}}$ &Text &Icon & Avg \\
    \midrule
     & \ding{55}  & \ding{55}    &53.7\% &15.7\% & 39.2\%     \\ 
     & \ding{51}  & \ding{55}    &55.2\% &16.7\% & 40.5\%     \\ 
     & \ding{55}  & \ding{51}    &53.8\% &18.2\%  & 40.2\%   \\ 
     & \ding{51}  & \ding{51}    &58.0\% &17.4\%  & 42.5\%   \\ 
    \bottomrule
\end{tabular}
\caption{
Ablation study on the effect of different reward function combinations.
$\mathcal{R}_{\text{div}}$ denotes the IoU-based reward for diversity filtering, and $\mathcal{R}_{\text{acc}}$ denotes the Monte Carlo–based reward for accuracy supervision.
When $\tau = 1$ and $\delta = 1$, all preference pairs are retained for training. 
% Varying either hyperparameter effectively controls the proportion of filtered samples.
}
\label{tab:filter_comparison}
\end{table}

\begin{figure}[t]
\centering
\includegraphics[width=0.9\columnwidth]{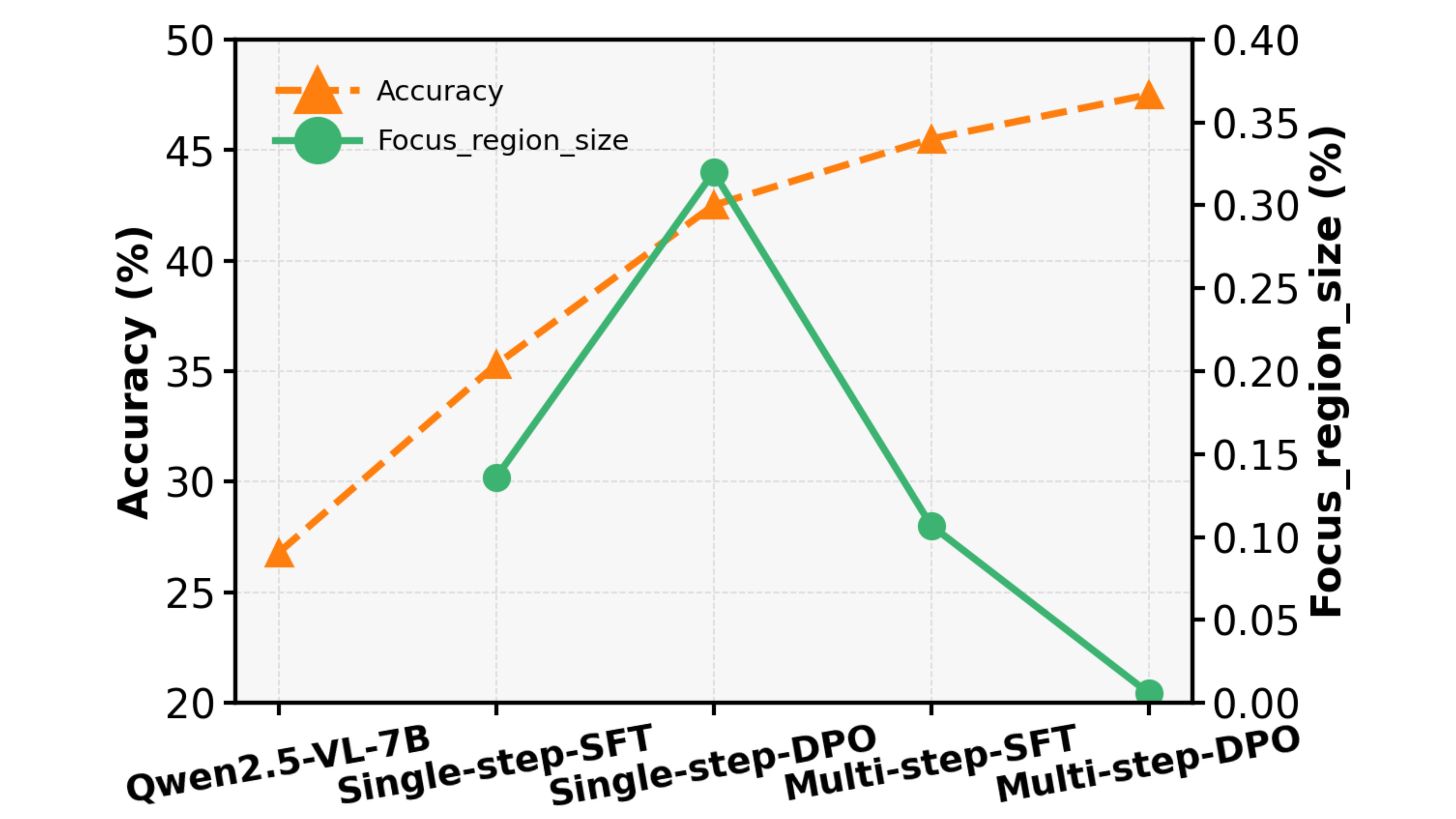} 
\caption{
Comparison of model accuracy across training stages and the corresponding change in final-step focus region size during inference.
}
\label{fig:Perfermance_increasing_along_with_stage}
\end{figure}

\begin{figure}[t]
\centering
\includegraphics[width=0.9\columnwidth]{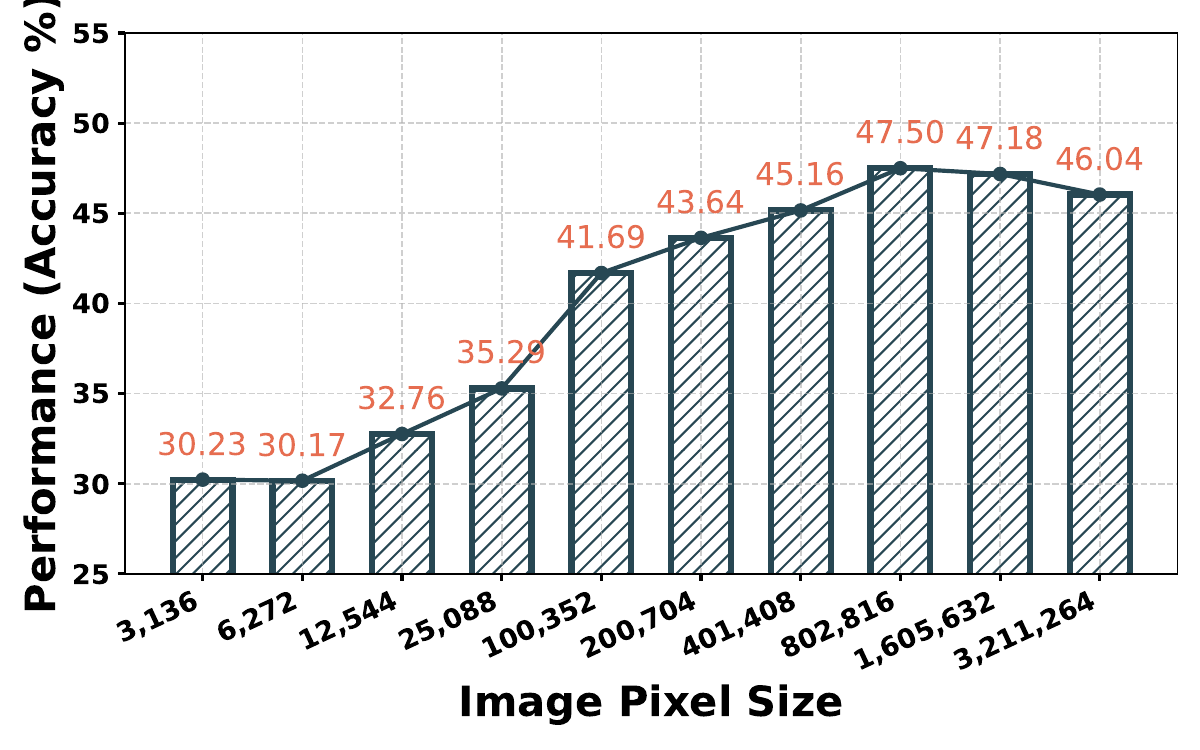}
\caption{
Ablation study on the effect of the \texttt{min-pixels} threshold during inference on the ScreenSpot-Pro benchmark, by varying its value from $3136$ to $3{,}211{,}264$.
% Ablation study on the effect of \texttt{min-pixels} during inference.
% We fix the \texttt{max-pixels} to $4816896$ and vary the \texttt{min-pixels} threshold from $3136$ to $3211264$.
}
\label{fig:abalation_min_pixels_affect}
\end{figure}

\begin{figure*}[t]
\centering
\includegraphics[width=2\columnwidth]{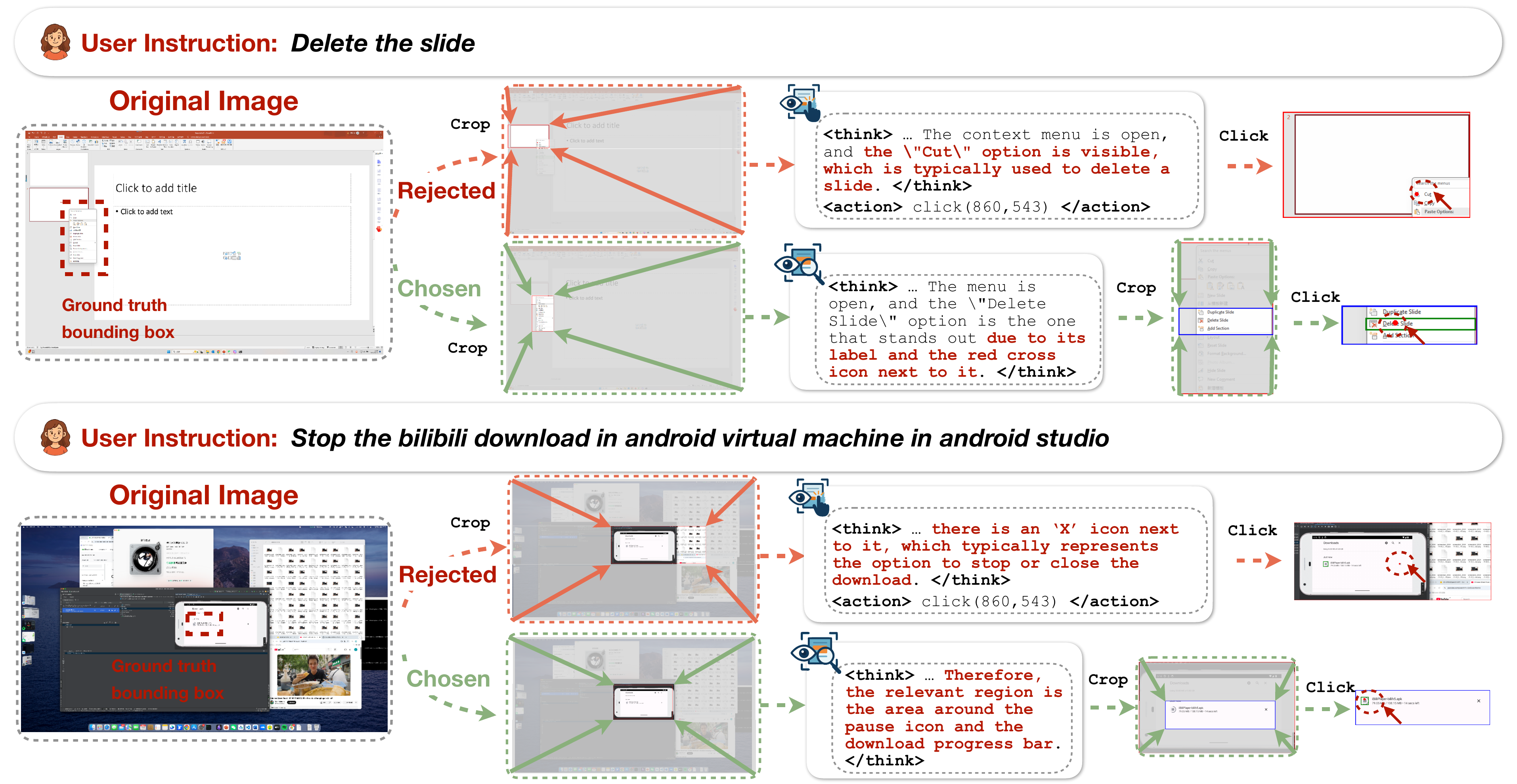}

\caption{
Comparison of single-step and multi-step perception. 
In both examples, the single-step model makes mistakes: either picking the wrong region or being misled by confusing elements, such as mistakenly selecting the “Cancel Download” icon instead of the intended “Pause Download” button. 
The multi-step model first zooms in on a rough area and then refines its choice, successfully finding the target. 
% This shows how multi-step perception makes the model more accurate and reliable for complex GUI tasks.
}
\label{fig:two_crop_vs_one_crop_case_analysis}
\end{figure*}

%%%%%%%%%%%% DPO 训练数据构造时的两个过滤规则的消融实验
\subsection{Ablation of Quality Estimation}
To assess the contribution of each reward scoring function introduced in Section~\ref{sec:region_wise_preference_learning}, we perform an ablation study covering all possible combinations of these signals. We fix $\delta = 4$ to ensure high confidence in the correctness of preference labels, and set $\tau = 0.8$ to promote diversity in the selected pairs.
As summarized in Table~\ref{tab:filter_comparison}, the removal of any individual scoring function results in a noticeable drop in performance, underscoring their complementary effects. These signals jointly enable effective filtering of low-quality samples, thereby improving the robustness and generalization of preference-based training.

\subsection{The Effect of Self-Evolving Framework}
We conduct a detailed analysis of how the self-evolving learning algorithm improves model performance over training stages. 
As shown in Figure~\ref{fig:Perfermance_increasing_along_with_stage}, each stage contributes to consistent performance gains, demonstrating the effectiveness of progressive self-supervised learning. 
In addition, we track the size of the focus region right before decision making. 
The results show that the final model is able to make accurate predictions using only ~20\% of the original image, which aligns well with our findings in the preliminary study.

% mix-pixels对 模型推理性能的影响
\subsection{The Effect of \texttt{min-pixels} During Inference}
To investigate how the minimum image resolution affects model performance, we conduct an ablation study on the \texttt{min-pixels} parameter.
% This parameter defines a lower bound on the number of pixels of images. 
If the pixel count of an image falls below \texttt{min-pixels}, the image is automatically resized to meet this threshold.

As shown in Figure~\ref{fig:abalation_min_pixels_affect}, performance drops significantly when \texttt{min-pixels} is too small. 
As the threshold increases, the model benefits from enhanced visual input, leading to performance improvements. 

These findings highlight the importance of carefully tuning the \texttt{min-pixels} value to ensure sufficient visual detail in cropped regions, while avoiding excessive upsampling that may harm accuracy in multi-step perception.

\begin{figure}[t]
\centering
\includegraphics[width=0.45\textwidth]{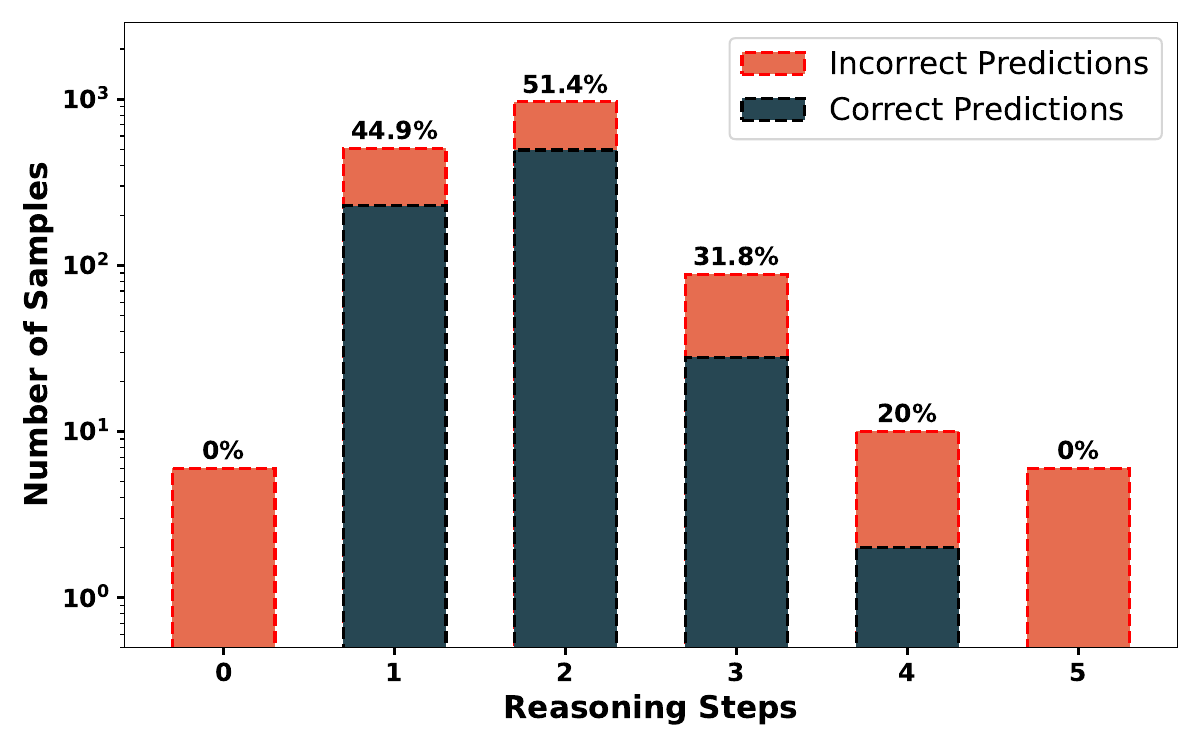} 
\caption{Proportion of correct and incorrect predictions made by LASER at different reasoning steps.}
\label{fig:ablation_reasoning_step}
\end{figure}

\subsection{Reasoning Step Analysis}
~\label{sec:reasoning_step}
We analyze the proportion of correct and incorrect predictions made by LASER across different reasoning steps.
As shown in Figure~\ref{fig:ablation_reasoning_step}, LASER achieves peak accuracy at two-step reasoning.
Interestingly, despite being trained only on two-step trajectories, the model adaptively perform multi-step reasoning during inference, showcasing its adaptive reasoning capabilities.

%%%%% Analyse Case %%%%%
\subsection{Analysis of Typical Cases}
To further understand the effectiveness of the multi-step perception, we conduct a qualitative comparison between the single-step and the multi-step perception approach. As shown in Figure~\ref{fig:two_crop_vs_one_crop_case_analysis}, the multi-step perception strategy enables the model to narrow down its attention to more fine-grained and semantically relevant regions.
Specifically, in many challenging cases where the target region is either small, visually ambiguous, or embedded within dense GUI components, the single-step crop often fails to localize accurately due to the overwhelming global context. In contrast, our two-step crop first identifies a coarse candidate area and then refines the prediction within this region, effectively filtering out irrelevant areas and focusing on semantically meaningful sub-regions.

%%%%%%%%%%%%%%%%%%%%%%%%%% Conclusions %%%%%%%%%%%%%%%%%%%%%%%%%%%%%%%%%%%%%%%%%%%%%%%%%%
\section{Conclusion}
In this paper, we introduce \textbf{LASER}, a self-evolving training strategy designed to elicit active perception capability of VLMs in GUI grounding tasks. Unlike prior methods that rely on static perception or strong supervision, LASER progressively guides the model to perform multi-step active visual perception and coordinate prediction. Through a combination of multi-stage optimization, Monte Carlo reward estimation, and IoU-based trajectory filtering, our method encourages active perception and self-improvement without human-annotated trajectories or teacher models. Experimental results on ScreenSpot and ScreenSpot-Pro benchmarks demonstrate that LASER significantly improves grounding accuracy and generalization, particularly in high-resolution and complex multi-element GUI scenarios.

%%%%%%%%%%%%%%%%%%%%%%%%%%%%%%%%%%%%%%%%%%%%%%%%%%%%%%%%%%%%%%%%%%%%%%%%%%%%%%%%%%%%%%%%%%%%%%%%%%%%%%%%
\bibliography{aaai25}

\end{document}